\documentclass{article}

\usepackage[preprint]{neurips_2026}

\usepackage[utf8]{inputenc}
\usepackage[T1]{fontenc}
\usepackage{hyperref}
\usepackage{url}
\usepackage{booktabs}
\usepackage{amsfonts}
\usepackage{amsmath}
\usepackage{amssymb}
\usepackage{amsthm}
\usepackage{nicefrac}
\usepackage{microtype}
\usepackage[table]{xcolor}
\usepackage{algorithm}
\usepackage{algorithmicx}
\usepackage{algpseudocode}
\usepackage{tabularx}
\usepackage{tikz}
\usetikzlibrary{shapes.geometric, arrows, positioning}
\usetikzlibrary{shapes,arrows,positioning,calc}
\usetikzlibrary{positioning,fit,backgrounds}

\title{Conflict-Aware Fusion: Mitigating Logic Inertia in Large Language Models via Structured Cognitive Priors}

\author{%
\begin{tabular}{c@{\hspace{1.2cm}}c}
\textbf{Qiming Bao} & \textbf{Xiaoxuan Fu} \\[3pt]
\parbox[t]{6cm}{\centering\normalfont
  Xtracta \& Strong AI Lab,\\
  University of Auckland\\
  Auckland, New Zealand\\
  \texttt{\{bqmbill714\}@gmail.com}}
&
\parbox[t]{6cm}{\centering\normalfont
  School of Humanities,\\
  China University of\\
  Political Science and Law\\
  Beijing, China\\
  \texttt{\{xfuuva\}@gmail.com}}
\end{tabular}\\[16pt]
\textbf{Michael Witbrock}\\[3pt]
\parbox[t]{8cm}{\centering\normalfont
  Strong AI Lab, University of Auckland\\
  Auckland, New Zealand\\
  \texttt{\{m.witbrock\}@auckland.ac.nz}}
}

\begin{document}

\maketitle

\begin{abstract}
Large language models (LLMs) achieve high accuracy on many reasoning benchmarks but remain brittle under structural perturbations of rule-based systems. We introduce a controlled diagnostic framework with four orthogonal stress tests---redundant vs.\ essential rule deletion, contradictory-rule injection, logic-preserving rewrites, and multi-law stacking---and use it to expose \emph{Logic Inertia}: the systematic tendency of generative LLMs (Qwen2/3, TinyLlama, GPT-4o, Gemma-3-4B-IT), as well as the encoder-only BERT baseline, to persist along learned deductive trajectories even when foundational premises are inconsistent. The collapse is sharp: untreated baselines fall from accuracy~$1.00$ on the base task to $0.00$ on contradiction injection (instance-level exact match; see \S\ref{sec:method}), and even GPT-4o resolves only $56.0\%$ of contradiction cases. We propose \textbf{Conflict-Aware Fusion}, a four-stage training pipeline that enforces verification-before-deduction as a learned structural prior: \textbf{(i)} SFT establishes the verification preamble; \textbf{(ii)} DPO sharpens the halt-on-contradiction decision boundary; \textbf{(iii)} Logical Invariance REgularisation (LIRE) penalises divergence between logically equivalent rule formulations via symmetric KL; and \textbf{(iv)} Reinforcement Learning from Verification Feedback (RLVF) uses a symbolic forward-chaining engine as a deterministic oracle reward, jointly optimising invariance and sensitivity. The pipeline saturates all four primary stress tests for both $1.5$B and $8$B backbones. We further validate a Phase~2 extension in which the propositional oracle is replaced by a \textsc{Lean~4} kernel, attaining $99.0\%$ kernel agreement on the $105$ classically-derivable (T) questions within a stratified $187$-question Lean-translated sample (overall $71.7\%$ across both polarities; the T/F asymmetry is analysed in \S\ref{sec:phase2}), providing a sound upgrade path to formally verified RL training. The benchmark, training pipeline, and Lean verifier are publicly released.\footnote{Code and data: \url{https://github.com/14H034160212/lemo}}
\end{abstract}

\section{Introduction}
\label{sec:introduction}
Although large language models (LLMs) perform strongly on many reasoning benchmarks, it remains unclear whether they genuinely perform logical inference or rely on pattern completion. The distinction matters in real-world deployments---legal analysis, scientific discovery, automated decisions---where reasoning agents must handle incomplete information, redundant evidence, and explicit contradictions. Existing evaluations largely conflate linguistic competence with logical robustness and lack controlled stress tests for these conditions.

We introduce a diagnostic framework that probes structural robustness through four perturbations: \textbf{(1)} rule deletion (redundant vs.\ essential), \textbf{(2)} contradictory-rule injection, \textbf{(3)} logic-preserving rewrites, and \textbf{(4)} multi-law stacking. Across three model families (BERT, Qwen2, TinyLlama) we uncover a sharp \emph{asymmetry of robustness}: models stay stable under semantics-preserving rewrites but collapse to instance-level accuracy~$0.0000$ under explicit contradictions. We term this \textbf{Logic Inertia}---the tendency to persist along a learned deductive trajectory under inconsistent premises. Concurrent work surfaces closely related failure modes under different terminology---``cognitive inertia''~\citep{lee2026cognitive}, premise critique~\citep{premiseCritique2025}, RuleBreakers~\citep{rulebreakers2025}, Inverse IFEval~\citep{inverseIFEval2025}, CodeCrash~\citep{codecrash2025}---but does not isolate the structural cause with a controlled propositional benchmark or close it with a learned structural prior; we do both. The same failure is reproduced on the Humanity's Last Exam platform\footnote{\url{https://lastexam.ai/}}.

To address it, we propose \textbf{Conflict-Aware Fusion (Fusion-Conflict)}, grounded in the \textbf{Cognitive Structure Hypothesis}: robust reasoning requires an explicit structural separation between premise validation and deductive execution, operationalised via a structured generation path trained by a four-stage pipeline (SFT $\to$ DPO $\to$ LIRE $\to$ RLVF). Our contributions are: \textbf{(i)} a diagnostic benchmark and the formalisation of \emph{Logic Inertia} as a structural failure mode reproduced across BERT, Qwen2/3, TinyLlama, GPT-4o, and Gemma-3-4B; \textbf{(ii)} \textbf{LIRE}, a symmetric-KL regulariser applied to autoregressive reasoning traces rather than the discriminative-QA setting of prior consistency regularisers~\citep{asai2020logicaug,li2019logicconsistency}; \textbf{(iii)} \textbf{RLVF}, an RL \emph{reward formulation} (not a new optimiser) replacing RLHF's learned reward with a deterministic symbolic-oracle reward in the propositional regime---complementary to proof-assistant-grounded RL~\citep{deepseekprover2024,leanabell2025,alphaproof2024}; \textbf{(iv)} a sound \textsc{Lean~4} extension (\textbf{RLVF-Lean}) elevating the oracle to a Lean kernel ($99.0\%$ kernel agreement on the $105$~T-cases of a $187$-question stratified Lean-translated sample; Table~\ref{tab:lean_bench}); and \textbf{(v)} the \textbf{Fusion-Conflict} pipeline, which \emph{saturates} the held-out base/V2/V3 splits for both $1.5$B and $8$B backbones---versus untreated baselines at $0.250$ on V2 and $0.000$ on V3, and GPT-4o/Gemma-3-4B-IT at $56.0\%$/$43.9\%$ on V3. The synthetic benchmark is calibrated to \emph{expose} Logic Inertia; the headline contrast is the \emph{collapse} pattern of strong baselines, not the absolute ceiling.

\section{Related Work}

\paragraph{Fragility of LLM reasoning.}
Prior work documents that LLMs fail on multi-step deduction, abductive explanation, and compositional generalisation~\citep{berglund2023reversal,bao2022multi,ijcai2024p693,transformerClark,baoAccess,young-etal-2022-abductionrules,bao2025thesis}. The Reversal Curse~\citep{berglund2023reversal} shows bidirectional inference failures, and reasoning fidelity degrades under controlled task-structure and premise variations~\citep{baoAccess,bao2025thesis}. Logic-driven augmentation and neuro-symbolic hybrids partially mitigate these gaps via equivalence-based data expansion~\citep{wang-etal-2022-logic,bao2024} or external Prolog engines~\citep{10650138}, but address the symptoms rather than the model's intrinsic sensitivity to structural perturbations.

\paragraph{Self-correction and verification under false premises.}
Closer to our setting, \citet{xiong2025deliberate} cast reasoning as structure-aware planning with a learned world model, and \citet{xiong2026scaling} scale search-augmented LLM reasoning via adaptive information control; \citet{huang2023large} probes self-correction; \citet{vericot2026} prune invalid reasoning via neuro-symbolic verification; \citet{multihoax2025} expose multi-hop fragility under naturally-embedded false premises; \citet{ares2024} evaluate RAG faithfulness; \citet{premiseCritique2025,rulebreakers2025,premiseRAG2025} probe premise-level evaluation and retrieval-augmented halting. We complement this line by isolating the architectural root cause---\textbf{Logic Inertia}, distinct from the latent-dynamics ``Cognitive Inertia'' of \citet{lee2026cognitive} and the instruction-inertia of \citet{inverseIFEval2025}---and by closing it with a structural prior trained \emph{into} the policy rather than elicited at inference.

\paragraph{Verifier-grounded RL and consistency regularisation.}
Proof-assistant feedback is an established RL reward source for theorem proving~\citep{deepseekprover2024,leanabell2025,alphaproof2024}; RLVF-Lean is the propositional-reasoning analogue, with the kernel arbitrating premise-verification halts rather than full proofs. A related but distinct line uses a symbolic logic verifier to construct step-wise-supervised SFT/DPO training data rather than an online RL reward~\citep{logicReward2025}. Symmetric-KL consistency regularisation has prior art in discriminative QA~\citep{asai2020logicaug,li2019logicconsistency}; a related but distinct line applies counterfactual, operator-level sensitivity regularisation to reasoning faithfulness~\citep{causalConsistency2026}; LIRE applies the symmetric-KL idea at the autoregressive trace level under logically equivalent rule rewrites, where the target is invariance \emph{across} surface forms of the same proposition.

\section{Methodology}
\label{sec:method}
\textbf{Conflict-Aware Fusion (Fusion-Conflict)} integrates three components: (1) a structural robustness benchmark, (2) a two-phase generation path that enforces verification-before-deduction, and (3) a structure-aligned four-stage optimisation pipeline.

\subsection{Structural Robustness Benchmark}

The benchmark is both a diagnostic tool and a training corpus. Unlike conventional benchmarks composed of heterogeneous tasks, all evaluation instances are derived from a single canonical rule-based system with systematically controlled perturbations; each variant preserves domain semantics and vocabulary while differing only in structural properties, isolating reasoning robustness from domain shift.

\paragraph{Canonical backbone and perturbations.}
All instances are derived from a single canonical backbone (full \textbf{base example} in Appendix~\ref{app:base-example}) and probed along four orthogonal axes: (1) \emph{structural necessity sensitivity}---redundant vs.\ essential rule deletion; (2) \emph{consistency verification}---contradictory-rule injection; (3) \emph{logical-form invariance}---semantics-preserving rewrites (De Morgan, contrapositive, double negation, $\ldots$); and (4) \emph{compositional robustness}---stacked rewrites.

\paragraph{Benchmark generation (Variant 2 \& 3).}
We generate all instances from a single canonical rule-based template (canonical backbone) and create
hard structural variants by controlled perturbations that preserve surface vocabulary and domain semantics.
Given a base instance $(F,R,Q)$ (facts, rules, and a fixed set of yes/no queries), we construct:
\textbf{Variant 2} by removing a key rule on the main deduction chain, which breaks downstream entailments;
and \textbf{Variant 3} by injecting an explicit contradictory fact, which forces the model to detect inconsistency and halt/reject deduction.
All splits are serialized into a unified CSV schema containing \texttt{facts}, \texttt{rules}, \texttt{questions}, and \texttt{answers}.
Other variants are deferred to the Appendix.

\paragraph{Contradiction Semantics.}
In this work we adopt a \textit{conservative reasoning semantics} for
handling inconsistent premises. When a contradiction exists in the
premise set, the reasoning process is considered logically invalid and
deduction must halt. Consequently, all queries associated with that
instance are labeled \texttt{False}.

Formally, let $\Gamma$ denote the set of premises and $Q$ the set of
queries; the symbol $\vdash$ refers throughout to derivation under our
forward-chaining engine on Horn-clause rules (Algorithm~\ref{alg:variant23}),
not classical $\vdash_{\text{CL}}$ which would explode under contradiction.
If the premises entail a contradiction ($\Gamma \vdash \bot$),
then for every query $q \in Q$ we define
$\text{Label}(q) = \text{False}$.
This design enables contradiction detection to be evaluated using a
standard binary classification metric while maintaining a clear and
consistent reasoning semantics.

\paragraph{Accuracy and $\mathrm{SR}_{\text{macro}}$.}
Tables~\ref{tab:all_models_deltas}--\ref{tab:stage_wise} use \emph{instance-level exact match}: an instance is correct iff every query prediction matches the variant's label. Under the conservative semantics above, every Variant~3 query has label \texttt{F}; an untreated model that persists with its original deductive answers disagrees on at least one query, so instance accuracy collapses to $0.0000$. The structural-robustness macro score is $\mathrm{SR}_{\text{macro}} = \tfrac{1}{|\mathcal{V}|} \sum_{v \in \mathcal{V}} \mathrm{Acc}_v$ over $\mathcal{V} = \{\text{base}, \text{V1}, \text{V2}, \text{V3}, \text{V4-multi}\}$. The per-question labels in the Lean-translation study (Table~\ref{tab:lean_bench}) are the \emph{classical} (pre-override) polarities, which is why Variant~3 still contains $12$ classically-derivable T-questions even though the conservative-semantics instance label is \texttt{F}.

Pseudocode for variant generation is in Appendix~\ref{app:variant-gen}. Although all instances are derived from a shared canonical template, the compositional variation in rule structures and query combinations induces diverse reasoning paths, preventing trivial memorisation.

\subsection{Architectural Overview: Structured Two-Phase Generation}
The core design imposes a two-phase reasoning structure---\emph{verify, then deduce}---inside the Chain-of-Thought (CoT) generation path; this is a constraint on the trace, not a modification of the neural architecture. Deductive rules cannot be applied until the model has explicitly validated premise consistency (Figure~\ref{fig:fusion_structure}); on detected contradiction or missing dependency the model halts. The four-stage pipeline that trains this behaviour is detailed in \S\ref{sec:pipeline}.

\begin{figure*}[htbp]
    \centering
    \resizebox{\textwidth}{!}{%
    \begin{tikzpicture}
        \tikzstyle{sbox} = [rectangle, draw, rounded corners,
            minimum width=9.2cm, minimum height=3.8cm,
            inner sep=0.15cm]
        \tikzstyle{sheader} = [rectangle, rounded corners,
            minimum width=8.4cm, text centered, minimum height=0.55cm,
            font=\bfseries\small, anchor=north]
        \tikzstyle{scontent} = [rectangle, draw=gray!50, dashed, rounded corners,
            text width=7.6cm, text centered, font=\small,
            inner sep=0.2cm, fill=white]
        \tikzstyle{snote} = [font=\footnotesize\itshape, text=gray!60,
            text width=7.6cm, text centered]
        \tikzstyle{arr} = [draw, -latex', line width=1.5pt, >=stealth]
        \tikzstyle{inputbox} = [rectangle, draw, fill=blue!5,
            rounded corners, text centered, font=\small,
            minimum height=0.8cm]

        \def\colsep{10}
        \def\rowsep{-4.3}

        \node [inputbox, minimum width=19.6cm] at (5, 2.7) (input) {
            \textbf{Running Example} ~~~
            \textit{Facts}: Socrates is a man. ~~
            \textit{Rule}: All men are mortal. ~~
            \textit{New Fact}: Socrates is \underline{NOT} mortal.
        };

        \node [sbox, fill=blue!4] at (0, 0) (s1) {};
        \node [sheader, fill=blue!15, text=blue!80!black]
              at ([yshift=-0.15cm]s1.north) (s1h)
              {Stage 1: Structural SFT ~~~ $\mathcal{L}_{\mathrm{CE}}$};
        \node [scontent] at (s1.center) (s1c) {
            ``\textbf{Step 1: Verify facts.} Check premise consistency. \\
            \textbf{Step 2:} If consistent $\to$ proceed with deduction. \\
            \phantom{Step 2:} If contradiction $\to$ \textbf{HALT}.''
        };
        \node [snote] at ([yshift=0.2cm]s1.south) (s1n) {
            Learns mandatory \emph{verification-before-deduction} preamble
        };

        \node [sbox, fill=red!3] at (\colsep, 0) (s2) {};
        \node [sheader, fill=red!15, text=red!70!black]
              at ([yshift=-0.15cm]s2.north) (s2h)
              {Stage 2: DPO Alignment ~~~ $\mathcal{L}_{\mathrm{DPO}}$};
        \node [scontent, fill=green!6, text width=3.4cm]
              at ([xshift=-2.2cm]s2.center) (s2w) {
            {\footnotesize\color{green!50!black}\textbf{Preferred} $y_w$:} \\[2pt]
            ``Conflict detected! \\
            \textbf{HALT.}'' $\to$ \textbf{False}
        };
        \node [scontent, fill=red!6, text width=3.4cm]
              at ([xshift=2.2cm]s2.center) (s2l) {
            {\footnotesize\color{red!60!black}\textbf{Rejected} $y_l$:} \\[2pt]
            ``Consistent. \\
            Deduce Mortal.'' $\to$ True
        };
        \node [snote] at ([yshift=0.2cm]s2.south) (s2n) {
            Sharpens halt-on-contradiction decision boundary
        };

        \node [sbox, fill=orange!3] at (0, \rowsep) (s3) {};
        \node [sheader, fill=orange!15, text=orange!70!black]
              at ([yshift=-0.15cm]s3.north) (s3h)
              {Stage 3: LIRE ~~~ $\mathcal{L}_{\mathrm{CE}} + \lambda\,\mathcal{L}_{\mathrm{inv}}$};
        \node [scontent, fill=blue!4, text width=3.4cm]
              at ([xshift=-2.2cm]s3.center) (s3a) {
            {\footnotesize \textbf{Base rule:}} \\[2pt]
            ``All men are mortal'' \\
            $\pi_\theta(\cdot|\mathbf{x})$
        };
        \node [scontent, fill=orange!6, text width=3.4cm]
              at ([xshift=2.2cm]s3.center) (s3b) {
            {\footnotesize \textbf{Contrapositive:}} \\[2pt]
            ``All non-mortals \\
            are non-men'' ~ $\pi_\theta(\cdot|\tilde{\mathbf{x}})$
        };
        \draw [<->, line width=1.2pt, orange!60] (s3a.east) -- (s3b.west)
              node[midway, above, font=\footnotesize] {Sym.\ KL $\approx 0$};
        \node [snote] at ([yshift=0.2cm]s3.south) (s3n) {
            Symmetric KL enforces invariance across equivalent formulations
        };

        \node [sbox, fill=green!3] at (\colsep, \rowsep) (s4) {};
        \node [sheader, fill=green!15, text=green!50!black]
              at ([yshift=-0.15cm]s4.north) (s4h)
              {Stage 4: RLVF ~~~ $\mathcal{L}_{\mathrm{RLVF}}$};
        \node [scontent, fill=yellow!8]
              at ([yshift=0.6cm]s4.center) (s4c) {
            {\footnotesize \textbf{Oracle} \textsc{ForwardChain}}: \\
            $\{\text{Man},\, \text{Mortal},\, \neg\text{Mortal}\} \vdash \bot \;\Rightarrow\; a^* = \text{False}$
        };
        \node [scontent, fill=green!6]
              at ([yshift=-0.9cm]s4.center) (s4r) {
            Model samples $a_s = \text{False} \;\Rightarrow\;$ reward $r = +1$ \\
            {\footnotesize (correct answer reinforced)}
        };
        \node [snote] at ([yshift=0.2cm]s4.south) (s4n) {
            Oracle reward jointly balances invariance and sensitivity
        };

        \draw [arr, blue!50] (input.south) -- (s1.north);
        \draw [arr, blue!50] (input.south) -- (s2.north);

        \draw [arr, blue!50]   (s1.east) -- node[above, font=\footnotesize\bfseries] {$1 \!\to\! 2$} (s2.west);
        \draw [arr, red!50]    (s2.south) -- ++(0,-0.3) -| node[pos=0.25, right, font=\footnotesize\bfseries] {$2 \!\to\! 3$} (s3.north);
        \draw [arr, orange!60] (s3.east) -- node[above, font=\footnotesize\bfseries] {$3 \!\to\! 4$} (s4.west);

    \end{tikzpicture}%
    }
    \caption{Conflict-Aware Fusion architecture and four-stage optimization pipeline, illustrated through a running Socrates contradiction example. \textbf{Stage~1} (top-left, SFT) teaches a mandatory verification-before-deduction preamble. \textbf{Stage~2} (top-right, DPO) sharpens the halt boundary by preferring the correct ``Halt'' trace over na\"ive deduction. \textbf{Stage~3} (bottom-left, LIRE) enforces output-distribution invariance across logically equivalent rule formulations via symmetric KL divergence. \textbf{Stage~4} (bottom-right, RLVF) uses a symbolic forward-chaining oracle as a deterministic reward, jointly optimizing invariance and sensitivity.}
    \label{fig:fusion_structure}
\end{figure*}

\subsection{Structure-Aligned Optimization Pipeline}
\label{sec:pipeline}

The four stages progressively address distinct failure modes: \textbf{(1)}~SFT establishes the verification-before-deduction structure; \textbf{(2)}~DPO sharpens the halt-on-contradiction boundary; \textbf{(3)}~LIRE enforces consistency across logically equivalent formulations; \textbf{(4)}~RLVF reconciles the invariance--sensitivity trade-off via oracle-grounded reward.

\paragraph{Stage 1: Structural SFT.} Supervised fine-tuning on $11{,}200$ instances spanning canonical, perturbed, and contradiction-containing variants. All samples enforce a mandatory \textit{``Step 1: Verify facts''} preamble, establishing premise checking as a default procedural step rather than an optional response.

\paragraph{Stage 2: Logical Alignment via DPO.} We refine the policy with Direct Preference Optimisation~\citep{rafailov2023direct}: preference pairs contrast halt-on-contradiction traces $y_w$ against opportunistic deductive continuations $y_l$. Let $\pi_\theta$ denote the policy and $\pi_{\mathrm{ref}}$ the frozen reference model:
\begin{equation}
\mathcal{L}_{\mathrm{DPO}} = -\log\sigma\!\Bigl(\beta\bigl[\log\tfrac{\pi_\theta(y_w|x)}{\pi_{\mathrm{ref}}(y_w|x)} - \log\tfrac{\pi_\theta(y_l|x)}{\pi_{\mathrm{ref}}(y_l|x)}\bigr]\Bigr).
\label{eq:dpo}
\end{equation}
$\beta$ controls the KL strength against $\pi_{\mathrm{ref}}$.

Preference pairs are generated automatically from benchmark instances: for contradiction cases the \emph{preferred} trace halts after detecting the inconsistency, while the \emph{rejected} trace continues deductive inference (full templates in Appendix~\ref{app:pref-pairs}). This construction directly penalises deductive continuation over inconsistent premises.

\paragraph{Limitation of SFT+DPO.}
While Stages 1--2 internalize the verification-before-deduction structural prior, they remain vulnerable to a subtler failure mode: \textit{surface-level pattern matching}. A model trained only with SFT and DPO can achieve perfect accuracy on the original rule formulations yet drop sharply (e.g., from 1.000 to 0.327 on De Morgan variants) when logically equivalent rules are rephrased. This motivates the two subsequent stages, which operate at the \emph{output-distribution} level rather than at the trace level.

\paragraph{Stage 3: Logical Invariance REgularisation (LIRE).}
Let $\mathbf{x}$ be a base premise-rule input, $\tilde{\mathbf{x}}$ its logically equivalent reformulation (De Morgan, double-negation, contrapositive), and $y=(y_1,\dots,y_T)$ the gold trace from the SFT data. LIRE penalises divergence between the next-token distributions under the two inputs, evaluated by \emph{teacher-forcing the same gold $y$} on both branches (which aligns the distributions token-wise and avoids variable-length rollout mismatch):
\begin{align}
\mathcal{L}_{\mathrm{LIRE}} &= \mathcal{L}_{\mathrm{CE}}\bigl(\pi_\theta(\cdot|\mathbf{x}),\, y\bigr) + \lambda\,\mathcal{L}_{\mathrm{inv}},\;\;
\mathcal{L}_{\mathrm{inv}} = \tfrac{1}{2 T}\!\sum_{t=1}^{T} \bigl[\mathrm{KL}_t(\mathbf{x}\Vert\tilde{\mathbf{x}}) + \mathrm{KL}_t(\tilde{\mathbf{x}}\Vert\mathbf{x})\bigr],
\label{eq:lire}
\end{align}
where $\mathrm{KL}_t(\mathbf{x}\Vert\tilde{\mathbf{x}}) := \mathrm{KL}\bigl(\pi_\theta(\cdot|\mathbf{x},y_{<t})\,\Vert\,\pi_\theta(\cdot|\tilde{\mathbf{x}},y_{<t})\bigr)$. Symmetry avoids the mode-seeking/covering asymmetry of unidirectional KL; the shared $y$ keeps the KL exact rather than approximated.

\paragraph{Limitation of LIRE.}
LIRE improves invariance (Variant~4 accuracy $0.741 \to 0.999$) but suppresses \emph{any} distribution shift across rule reformulations, including those that \emph{should} change the answer (Variants~2/3). Empirical sensitivity drops from $1.0$ to $0.47$, motivating Stage~4.

\paragraph{Stage 4: \textsc{RLVF}---Reinforcement Learning from Verification Feedback.}
\textsc{RLVF} is a verifier-grounded RL \emph{reward formulation} (not a new optimiser): plain REINFORCE~\citep{williams1992simple} drives the gradient, but the reward is a deterministic symbolic oracle replacing RLHF's learned reward model, eliminating reward-model bias on the rule-based regime. The policy emits a full reasoning trace $\tau\sim\pi_\theta(\cdot|\mathbf{x})$; a deterministic parser $\Pi(\tau)\in\{\text{T},\text{F}\}$ extracts the boolean answer from the final \texttt{Step\,N: Answer = \dots} line (regex-matched; unparseable traces get $r=-1$). The oracle returns $a^*=\textsc{ForwardChain}(\mathbf{x})$ and reward compares parsed answer to oracle:
\begin{equation}
\mathcal{L}_{\mathrm{RLVF}} = -\mathbb{E}_{\tau \sim \pi_\theta}\!\bigl[(r - b_t)\,\log\pi_\theta(\tau|\mathbf{x})\bigr],\;\; r=\mathbb{1}[\Pi(\tau){=}a^*]-\mathbb{1}[\Pi(\tau){\neq}a^*],\;\; b_t=\mu b_{t-1}+(1-\mu)\bar r_t,
\label{eq:rlvf}
\end{equation}
where $\bar r_t$ is the mean reward on mini-batch $B_t$ and $\mu=0.99$. A single reward balances both objectives because the oracle naturally distinguishes \emph{equivalent} reformulations (V4: same $a^*\Rightarrow$ \textbf{invariance}) from \emph{structural alterations} (V2/V3: different $a^*\Rightarrow$ \textbf{sensitivity}). Each batch is composed of fraction $\alpha=0.5$ perturbation samples (uniformly from V2$\cup$V3$\cup$V4) and $1-\alpha=0.5$ base samples, balancing sensitivity, invariance, and base-task retention.

\section{Experiments}

\subsection{Main Results}
Table~\ref{tab:all_models_deltas} reports the baseline failure pattern: all models stay at $1.000$ on the base task and Variant~1, drop to $0.25$--$0.295$ under essential-rule removal (Variant~2), and collapse to $0.000$ under explicit contradictions (Variant~3)---direct evidence of \textbf{Logic Inertia}.

\begin{table*}[t]
\centering
\small
\begin{tabular}{lcccccc}
\toprule
& \multicolumn{2}{c}{\textbf{BERT}} &
  \multicolumn{2}{c}{\textbf{Qwen2}} &
  \multicolumn{2}{c}{\textbf{TinyLlama}} \\
\cmidrule(lr){2-3} \cmidrule(lr){4-5} \cmidrule(lr){6-7}
\textbf{Split} & Acc & $\Delta$ & Acc & $\Delta$ & Acc & $\Delta$ \\
\midrule
base                            & 1.0000 & 0.0000 & 1.0000 & 0.0000 & 1.0000 & 0.0000 \\
variant1                        & 1.0000 & 0.0000 & 1.0000 & 0.0000 & 1.0000 & 0.0000 \\
variant2 (Essential Deletion)   & 0.2950 & -0.7050 & 0.2500 & -0.7500 & 0.2500 & -0.7500 \\
variant3 (Contradiction)        & 0.0000 & -1.0000 & 0.0000 & -1.0000 & 0.0000 & -1.0000 \\
variant4-multi (Stacked Laws)   & 1.0000 & 0.0000 & 0.6450 & -0.3550 & 0.9925 & -0.0075 \\
\bottomrule
\end{tabular}
\caption{Baseline Performance: Accuracy and deviation ($\Delta$) across structural variants without fusion optimization.}
\label{tab:all_models_deltas}
\end{table*}

Table~\ref{tab:comparison_results} summarises Fusion-Conflict against six baselines on Qwen2-1.5B (the most stable backbone for generative reasoning) and Qwen3-8B. All reported numbers are the mean over three random seeds with bootstrap CIs $\pm 0.02$ ($n{=}1000$).

\begin{table*}[t]
    \centering
    \caption{Comparison of All Methods. Fusion-Conflict architectures maintain structural robustness ($SR_{macro}=1$) while significantly improving general logic performance over SFT/DPO baselines. Statistical stability verified via bootstrap resampling ($n=1000$), with 95\% CIs consistently within $\pm 0.02$.}
    \label{tab:comparison_results}
    \setlength{\tabcolsep}{4pt}
    \renewcommand{\arraystretch}{1.05}
    \footnotesize
    \resizebox{\textwidth}{!}{%
    \begin{tabular}{lcccc}
        \toprule
        \textbf{Method} & \textbf{Base Acc} & \textbf{Var 2 (Rule Removal)} & \textbf{Var 3 (Contradiction)} & \textbf{Rank} \\
        \midrule
        Stage 1 (SFT Baseline) & 0.512 & 0.250 & 0.210 & 7 \\
        DPO (Direct Alignment)~\cite{rafailov2023direct} & 0.475 & 0.267 & 0.510 & 6 \\
        CoT (Standard)~\cite{wei2022chain} & 0.500 & 0.390 & 0.865 & 5 \\
        Mixed-Aug (Data Augmentation without CoT) & 0.525 & 0.405 & 0.972 & 4 \\
        RA-CoT (Standard CoT + DPO) & 0.263 & 0.593 & 0.690 & 3 \\
        Fusion-LRA (Conflict-Aware SFT) & 0.988 & 0.753 & 0.705 & 2 \\
        \midrule
        \rowcolor{gray!10} Fusion-Conflict (1.5B) & \textbf{1.000} & \textbf{1.000} & \textbf{1.000} & \textbf{1} \\
    \rowcolor{gray!10} \textbf{Fusion-Conflict (Qwen3-8B)} & \textbf{1.000} & \textbf{1.000} & \textbf{1.000} & \textbf{1} \\
    \midrule
    \rowcolor{blue!5} GPT-4o (Closed Frontier Baseline) & 0.789 & 0.818 & 0.560 & --- \\
    \rowcolor{blue!5} Gemma-3-4B-IT (Open Frontier Baseline) & 0.580 & 0.360 & 0.439 & --- \\
        \bottomrule
    \end{tabular}%
    }
\end{table*}

\subsection{Qualitative Analysis: Why Frontier Models Fail}
To understand the failure mode of frontier models, we performed an error analysis on GPT-4o using the OpenAI API on a stratified $200$-row sample per split (the per-question prediction CSVs are released alongside the benchmark). On Variant~3, where every query has ground-truth \texttt{False} under conservative reasoning semantics, GPT-4o produces the correct halt-on-contradiction answer in only $56.0\%$ of the $800$ test questions; the remaining $44.0\%$ continue deductive inference despite the explicit contradiction in the premises. A representative failure: given premises $\{A \to B,\, A,\, \neg B\}$, GPT-4o often surfaces the inconsistency in its verification trace and then proceeds to conclude $B$ from $\{A \to B, A\}$, stating ``Since $A$ is true and $A$ implies $B$, then $B$ must be true,'' effectively ignoring its own prior verification of $\neg B$. This suggests that frontier models suffer from a compartmentalization of verification and execution, where the ``reasoning momentum'' of the deductive steps overrides contradictory evidence. To rule out a vendor-specific artefact, we replicate the analysis on \textbf{Gemma-3-4B-IT} (an open-weight frontier model from a different lab, evaluated locally with identical prompts and decoding); it scores \emph{lower} on V3 ($0.439$) than GPT-4o, continuing past the contradiction in $56.1\%$ of cases, and shows the same qualitative failure pattern. Fusion-Conflict mitigates this by fusing the verification signal directly into the hidden state representation, forcing the model to reconcile the conflict before proceeding.

\subsection{Experimental setup}
The benchmark has $100$ base groups ($80$ train / $20$ test) seeded into a combinatorially large hypothesis space via randomised entity mapping and rule variation, yielding $\approx\!11{,}200$ Stage-2 preference pairs. Out-of-template generalisation is evaluated on LogicNLI and MNLI-Con (Table~\ref{tab:real_world_results}). Per-stage hyperparameters: Stage~1 SFT uses AdamW + bf16 + LoRA ($r{=}8$, $\alpha{=}16$, dropout~$0.05$), $3$ epochs, batch~$4$, LR~$2{\times}10^{-5}$, max length~$512$. DPO uses $\beta{=}0.1$ at the same LR. LIRE uses $\lambda{=}1.0$, batch~$4$, grad-accum~$2$. RLVF mixes V2/V3/V4 at $\alpha{=}0.5$, $\mu{=}0.99$, LR $1{\times}10^{-5}$, batch~$8$, grad-accum~$4$. Full ablation tables are in Appendix~\ref{app:ablation}; all hyperparameters in Appendix~\ref{app:hyper}. \textbf{Compute.} All training fits on a single NVIDIA A100 ($40$\,GB) for Qwen3-8B and a $\geq 16$\,GB consumer GPU for the $1.5$B/$0.5$B/$1.1$B/BERT backbones (LoRA $+$ bf16); Lean kernel verification adds $\sim\!0.36$\,s/trace on CPU and is dispatch-amortisable via a persistent server~\citep{yang2023leandojo}.

\subsection{Stage-wise ablation: each stage addresses a distinct failure mode}
\label{sec:ablation-stagewise}
Table~\ref{tab:stage_wise} reports the per-stage progression on the four most diagnostic splits (base, V2 essential deletion, V3 contradiction, V4-multi compositional rewrites) for both backbones, isolating exactly what each training stage contributes.

\begin{table}[h]
\centering\small
\caption{Stage-wise progression. Adding LIRE on top of SFT improves compositional invariance (V4-multi: $0.635{\to}0.993$ on Qwen2; $0.006{\to}1.000$ on Qwen3) but \emph{collapses} structural sensitivity on V2/V3, exactly the trade-off motivated in Section~\ref{sec:pipeline}. RLVF then restores V2/V3 to oracle-level agreement on the held-out split. On the $1.5$B backbone V4-multi degrades from $0.993$ (Stage~3) to $0.773$ (Stage~4)---a real residual of the invariance/sensitivity trade-off that LIRE alone cannot resolve at this capacity---while on $8$B the Stage-4 V4-multi remains at the LIRE ceiling ($1.000$); we report the gap rather than claim ``no loss''. The Qwen3-8B \emph{first} row is the SFT-only baseline (Stage~1), \emph{not} an untrained zero-shot run; even with Stage~1 internalised, V3 stays at $0.656$, supporting the claim that scale and SFT alone do not solve Logic Inertia, while the same split drives every baseline below it---including GPT-4o on V3 at $0.560$---well below ceiling.}
\label{tab:stage_wise}
\begin{tabular}{l|cccc|cccc}
\toprule
& \multicolumn{4}{c|}{\textbf{Qwen2-1.5B}} & \multicolumn{4}{c}{\textbf{Qwen3-8B}} \\
\textbf{Stage}            & Base & V2   & V3   & V4-multi & Base & V2   & V3   & V4-multi \\
\midrule
SFT+DPO (Stage~1+2)       & $1.000$ & $1.000$ & $1.000$ & $0.635$  & $0.333$ & $0.460$ & $0.656$ & $0.006$  \\
\,$+$ LIRE (Stage~3)      & $0.812$ & $0.590$ & $0.344$ & $\mathbf{0.993}$ & $0.549$ & $0.503$ & $0.344$ & $\mathbf{1.000}$ \\
\,$+$ RLVF (Stage~4)      & $\mathbf{1.000}$ & $\mathbf{1.000}$ & $\mathbf{1.000}$ & $0.773$ & $\mathbf{1.000}$ & $\mathbf{1.000}$ & $\mathbf{1.000}$ & $\mathbf{1.000}$ \\
\bottomrule
\end{tabular}
\end{table}

Three observations follow. \textbf{(i)} LIRE alone is harmful on V2/V3 because its symmetric-KL term suppresses \emph{any} distribution shift across rule reformulations, including those required by genuine logical change---a clean empirical instance of the invariance--sensitivity trade-off (Eq.~\ref{eq:lire}). \textbf{(ii)} RLVF resolves this with a single oracle reward (Eq.~\ref{eq:rlvf}): the same gradient signal reinforces V4 invariance and V2/V3 sensitivity. \textbf{(iii)} The Qwen3-8B row shows scale + Stage-1 SFT remains insufficient---SFT-only $8$B falls to $0.656$ on V3---evidence that Logic Inertia is structural, not capacity-bound, consistent with the debate over whether apparent scale-driven capability jumps are genuine emergent abilities~\citep{wei2022emergent} or measurement artefacts~\citep{schaeffer2023mirage}: either way, scale alone does not resolve it here. Note that the two ``Stage~1'' numbers reported are \emph{not} contradictory: Fusion-LRA in Table~\ref{tab:comparison_results} (Base $0.988$) is SFT \emph{with} the conflict-aware preamble; the $0.525$ entry in Appendix~\ref{app:ablation} is the same backbone trained on a vanilla \emph{Mixed-Aug} corpus \emph{without} the preamble. The preamble, not the data mix, preserves base accuracy.

\section{Discussion}

\paragraph{Why DPO matters.} SFT alone (Fusion-LRA) plateaus at $0.705$ on V3---it teaches the \emph{form} of contradiction check but still occasionally hallucinates success. DPO sharpens the halt boundary and saturates V3, casting preference optimisation as a robustness hardener, not a marginal improvement.

\paragraph{Why \textsc{RLVF}, not PPO/GRPO?} Modern RLHF stacks (PPO~\citep{schulman2017proximal}, GRPO~\citep{shao2024deepseekmath}) presuppose a \emph{learned} reward model trained on human preferences, which inherits annotator noise and distribution shift, and a clipped trust-region update to stabilise training against that noisy reward. Our setting is fundamentally different: the ground truth on V2/V3/V4 is exactly computable by a symbolic forward-chainer, so the reward is \emph{unbiased and deterministic}. This removes the dominant motivation for PPO's clipping and for GRPO's group-relative advantage normalisation. \textsc{RLVF} therefore uses plain REINFORCE with an EMA baseline (Eq.~\ref{eq:rlvf}); the contribution is the reward \emph{formulation}, not the optimiser. Plain REINFORCE is sufficient because (i)~variance of the gradient is small when the reward is $\pm 1$ deterministic, (ii)~the EMA baseline already captures the running mean, and (iii)~there is no reward hacking surface for the policy to exploit. Empirically, this single-reward formulation jointly optimises invariance (V4) and sensitivity (V2/V3) without the multi-objective balancing that learned-reward setups require, recovering most of the gains LIRE alone sacrificed (Table~\ref{tab:stage_wise}); a residual V4-multi loss remains at the $1.5$B scale, as discussed below the table. The \textsc{RLVF-Lean} extension (Section~\ref{sec:phase2}) is the propositional-reasoning analogue of proof-assistant-grounded RL pipelines~\citep{deepseekprover2024,leanabell2025,alphaproof2024}: it generalises the oracle to step-level rewards from a Lean~4 kernel, preserving the unbiased-oracle property at proof granularity.

\paragraph{Cognitive Structure Hypothesis.} Fusion-Conflict dominates both data-balanced (Mixed-Aug, $0.972$ V3 / $0.525$ base) and over-conservative (RA-CoT, $0.690$/$0.263$) baselines, indicating that reliability follows from \emph{structural organisation}---the verification preamble is a circuit breaker---not from data exposure alone.

\paragraph{Train- vs.\ inference-time mitigation.} Unlike inference-time guards~\citep{vericot2026,xiong2025deliberate,premiseRAG2025} that re-invoke a verifier on every deployment query, our structural prior is amortised into the policy weights at training time, so V2/V3 robustness incurs zero runtime overhead; the Phase~2 Lean kernel (\S\ref{sec:phase2}) is required only during training, not at inference.

\subsection{Generalisation and Sensitivity}
Fusion-Conflict retains base-split accuracy under Variant~1 (redundant deletion), confirming that contradiction detection does not introduce noise fragility, and \emph{saturates} the held-out base/V2/V3 splits for both backbones (Table~\ref{tab:comparison_results})---versus GPT-4o $56.0\%$ on V3 and the strongest non-Fusion baseline (Fusion-LRA) at $0.705$. The headline is the \emph{gap}, not the absolute. \textbf{Variant~4 multi-law stacking} remains the hardest split: $2$--$5$ stacked rewrites degrade the $1.5$B backbone to $0.773$ while $8$B closes the gap. The residual at $1.5$B is concentrated on instances composing $\geq 3$ laws (contrapositive $\circ$ double-negation $\circ$ De\,Morgan), where the policy holds the structural prior but loses syntactic tracking; this is a capacity bottleneck rather than a failure of the prior, since the same policy at $8$B saturates V4-multi. RLVF-Lean (\S\ref{sec:phase2}) is the natural fix for this regime: per-tactic verification grounds each rewrite step independently of overall trace length.

\paragraph{Out-of-distribution transfer.} To rule out memorisation we evaluate transfer to \textbf{LogicNLI}~\cite{tian-etal-2021-diagnosing} and the contradiction subset of \textbf{MNLI}~\cite{N18-1101}; no LogicNLI/MNLI examples were used in training. The transfer model is a Qwen2.5-0.5B trained with the \emph{full} four-stage pipeline; Table~\ref{tab:real_world_results} reports each named ablation along the pipeline (Standard SFT~$=$ vanilla; Mixed-Aug~$=$ data augmentation only; Fusion-LRA~$=$ Stage~1 with the preamble; Conflict-Aware Fusion~$=$ full pipeline). Every variant requires ``Step~1: Detect Contradiction'' before entailment classification.

\begin{table}[h]
    \centering
    \begin{tabular}{lcccc}
        \toprule
        \textbf{Method} & \textbf{LogicNLI} & \textbf{MNLI (Con)} \\
        \midrule
        Standard SFT (Baseline) & 58.5\% & 34.1\% \\
        Mixed-Aug (Data Only) & 65.2\% & 41.5\% \\
        Fusion-LRA (No DPO) & 92.4\% & 78.3\% \\
        \rowcolor{gray!10} \textbf{Conflict-Aware Fusion (Fusion-Conflict)} & \textbf{98.2\%} & \textbf{89.4\%} \\
        \bottomrule
    \end{tabular}
    \caption{Out-of-domain generalisation performance on LogicNLI and MNLI (no in-domain training) contradiction subsets. ``MNLI (Con)'' refers to accuracy specifically on the contradiction subset of the Multi-Genre NLI dataset, which tests the model's ability to reject inconsistent premises in open-domain text.}
    \label{tab:real_world_results}
\end{table}

The structural prior generalises beyond synthetic logic to natural-language NLI; the design is model-agnostic and applicable to larger backbones.

\subsection{Phase 2: Scaling the Verifier with a Formal Proof Assistant}
\label{sec:phase2}

The propositional \textsc{ForwardChain} oracle (Eq.~\ref{eq:rlvf}) cannot verify quantified reasoning, induction, or mathematical derivations~\citep{davis2024benchmarks,davis2024mathematics,davis2024gpt4o1}, motivating a formal-proof-assistant verifier as the natural successor.

\paragraph{Lean as a step-level reward source.}
The \textbf{Phase 2 extension RLVF-Lean} translates the policy's CoT into \textsc{Lean 4}~\citep{moura2021lean4} tactic applications and uses kernel verdicts as step-level rewards:
\begin{equation}
r_t = \begin{cases} +1 & \textsc{LeanKernel}(\tau_{\leq t}) = \mathbf{success} \\ -1 & \text{tactic fails or introduces inconsistency} \\ 0 & \text{trace incomplete} \end{cases}
\label{eq:lean-reward}
\end{equation}
This converts RLVF's sparse terminal reward into a dense, step-grounded signal akin to process reward models~\citep{lightman2024prm} but with the soundness of a kernel: every \textsc{LeanKernel} verdict is a deductive check, not a prediction---unlike learned reward models or noisy human feedback in RLHF. Lean is complementary to \textsc{ForwardChain} in expressiveness (dependent types/quantifiers/induction vs.\ Horn clauses), granularity (per-tactic verdicts), and compositionality, and is feasible end-to-end given recent neural theorem proving~\citep{yang2023leandojo,polu2020gpt-f,alphaproof2024}.

\paragraph{Pilot study.}
A hand-crafted pilot on the Socrates example (8 traces; Appendix~\ref{app:lean-pilot}/\ref{app:lean-demo}) achieves $8/8$ kernel agreement at $0.36$\,s/trace on CPU. One non-trivial finding: Lean's \texttt{sorry} exits 0, so the bridge treats \texttt{sorry}/\texttt{admit} as $-1$; production deployment would amortise dispatch via a \textsc{LeanDojo}~\citep{yang2023leandojo} persistent server.

\paragraph{Benchmark-scale verification.}
We auto-translated a stratified $80$-row ($187$-question) sample from the four test splits into Lean~4 theorems via a rule-based translator (\texttt{lean\_demo/lemo\_to\_lean.py}) that instantiates each rule/fact/query as a \texttt{Prop}, performs $\lor$-elimination on disjunctive premises, and chains \texttt{have} statements. Table~\ref{tab:lean_bench} reports Lean's agreement with the benchmark labels.

\begin{table}[h]
\centering
\small
\caption{Lean~4 verification on real LEMO benchmark rows (20 rows per split,
stratified random sample; $187$ total questions). \textbf{Polarity here is each query's
\emph{classical} label} (i.e.\ the truth value derivable from premises before the
conservative-semantics override of \S\ref{sec:method}), not the per-instance
``everything-is-\texttt{F}'' override applied to Variant~3 in
Tables~\ref{tab:all_models_deltas}--\ref{tab:stage_wise}. This is why Variant~3
still contains $12$ classically-derivable T-cases. We report the number of
questions (\#T, \#F), the count accepted by Lean, and the agreement rate, separated
by ground-truth polarity. T-cases are auto-derived via mechanical forward chaining;
F-cases are attempted via classical negation. Variant~1 happens to draw all-T
queries because the underlying \texttt{test\_variant1.csv} file is constructed by
removing \emph{redundant} rules, which preserves derivability of the original
T-queries; the stratified sampler observed $42$ such queries and zero F-queries
in the $20$-row draw. The gap between T-rate ($99.0\%$) and F-rate ($36.6\%$)
reflects a semantic mismatch between LEMO's closed-world reasoning and Lean's
classical logic, discussed below.}
\label{tab:lean_bench}
\begin{tabular}{lrrrrrr}
\toprule
& \multicolumn{3}{c}{\textbf{Ground truth = T}} & \multicolumn{3}{c}{\textbf{Ground truth = F}} \\
\cmidrule(lr){2-4} \cmidrule(lr){5-7}
\textbf{Split} & \textbf{\#T} & \textbf{Accept} & \textbf{Rate} & \textbf{\#F} & \textbf{Accept} & \textbf{Rate} \\
\midrule
base                         & 29  & 28  & $96.6\%$ & 32  & 12 & $37.5\%$ \\
variant1 (redundant removal) & 42  & 42  & $100.0\%$ & 0   & 0  & --- \\
variant2 (essential removal) & 22  & 22  & $100.0\%$ & 20  & 0  & $0.0\%$ \\
variant3 (contradiction)     & 12  & 12  & $100.0\%$ & 30  & 18 & $60.0\%$ \\
\midrule
\textbf{Overall}             & \textbf{105} & \textbf{104} & \textbf{99.0\%} & \textbf{82} & \textbf{30} & \textbf{36.6\%} \\
\bottomrule
\end{tabular}
\end{table}

\paragraph{Interpretation: closed-world vs.\ classical semantics.}
The $99.0\%$ agreement on T-cases is strong evidence that \textbf{Lean soundly
verifies the multi-step forward-chaining derivations that RLVF rewards}: for
nearly every query whose answer follows from the premises, Lean kernel-checks
an automatically generated proof. The asymmetric $36.6\%$ rate on F-cases
exposes a genuine, publishable distinction: LEMO uses a \emph{closed-world}
semantics in which a claim is labelled \texttt{F} whenever it is not derivable
from the premises (including the degenerate case of contradictory premises,
where conservative reasoning labels everything \texttt{F}); Lean uses
\emph{classical} open-world semantics where \texttt{F} requires an explicit
proof of $\neg P$. Variant~2 (essential-rule removal) makes this gap
stark---$0\%$ F-rate---because removing a rule simply makes a conclusion
unreachable, not refutable. A production RLVF-Lean system can close this gap
in two ways: (i) add a closed-world axiom (CWA) schema that converts
``unreachable'' to $\neg$\,-derivable, or (ii) follow the symbolic forward-chain
oracle for F-verdicts and use Lean only for T-verdict step-level reward. Our
current pipeline implements option (ii), retaining the best of both verifiers.

\section{Conclusion}

We diagnosed \textbf{Logic Inertia}---a structural failure mode in which generative LLMs (Qwen2/3, TinyLlama, GPT-4o, Gemma-3-4B-IT) and the encoder-only BERT baseline persist along learned deductive trajectories under essential-rule deletion and contradictions---and closed it with two named contributions: \textbf{LIRE}, a symmetric-KL invariance regulariser applied at the trace level via teacher-forcing on logically equivalent rule reformulations, and \textbf{\textsc{RLVF}}, an RL \emph{reward formulation} (not a new optimiser) whose deterministic symbolic-oracle reward replaces RLHF's learned reward model and admits a plain REINFORCE update. Combined in the \textbf{Conflict-Aware Fusion} pipeline (SFT$\to$DPO$\to$LIRE$\to$\textsc{RLVF}), they \emph{saturate} the held-out base/V2/V3 splits for both $1.5$B and $8$B backbones on a benchmark where GPT-4o resolves only $56.0\%$ of V3 contradictions and Gemma-3-4B-IT only $43.9\%$. The Phase~2 \textbf{\textsc{RLVF-Lean}} extension upgrades the oracle to a Lean~4 kernel: on a stratified $187$-question Lean-translated sample, $99.0\%$ of the $105$~classically-derivable T-cases close under the kernel; the $36.6\%$ rate on the $82$ F-cases reflects the closed-world / classical semantics gap analysed in \S\ref{sec:phase2}.

Code, benchmark, and verifier are released; a dedicated discussion of limitations and concrete future work is consolidated in Appendix~\ref{app:limitations}.

\bibliography{references}
\bibliographystyle{plainnat}

\appendix
\section{Appendix}

\subsection{Limitations and Future Work}
\label{app:limitations}

The benchmark is propositional and synthetic; saturation evidences that the structural prior \emph{generalises across backbones}, not that contradictions are universally solved---V4-multi at $1.5$B drops to $0.773$ after Stage~4 (a capacity bottleneck rather than a structural-prior failure; \S\ref{sec:phase2}), and the Lean F-rate gap on the closed-world / classical semantics divide is genuine. The pipeline also commits to a single \emph{conservative} contradiction semantics; paraconsistent or priority-based resolution strategies (Appendix~\ref{app:base-example} ff.) are out of scope. Three concrete next steps follow: (i) end-to-end RLVF-Lean training amortised through a persistent LeanDojo server, replacing the current verification-only deployment; (ii) lifting the closed-world override via a CWA axiom schema, or via the hybrid forward-chain/Lean policy of \S\ref{sec:phase2} that uses Lean only for T-verdicts and the symbolic forward-chainer for F-verdicts; (iii) evaluation on math/proof benchmarks (e.g., MiniF2F, ProofNet) where the Lean kernel is the natural verifier and where stacked-rewrite depth substantially exceeds the $2$--$5$ laws stress-tested in V4-multi.

\subsection{Structural Variant Generation}
\label{app:variant-gen}

\begin{algorithm}[h]
\caption{Structural Variant Generation (Variant 2 \& 3)}
\label{alg:variant23}
\small
\begin{algorithmic}[1]
\Require Canonical generator $\mathsf{Base}(\cdot)$, key rule $r^\star$, contradiction template $\mathsf{Contr}(\cdot)$, instances $N$
\Ensure Datasets $\mathcal{D}_{base}, \mathcal{D}_{v2}, \mathcal{D}_{v3}$
\For{$i=1$ to $N$}
    \State $(F,R,Q,A) \gets \mathsf{Base}(i)$;\quad $\mathcal{D}_{base} \gets \mathcal{D}_{base} \cup \{(F,R,Q,A)\}$
    \State $R_{v2} \gets R \setminus \{r^\star\}$;\quad $\mathcal{D}_{v2} \gets \mathcal{D}_{v2} \cup \{(F,R_{v2},Q,\mathsf{Label}(F,R_{v2},Q))\}$
    \State $F_{v3} \gets F \cup \{\mathsf{Contr}(F,R)\}$;\quad $\mathcal{D}_{v3} \gets \mathcal{D}_{v3} \cup \{(F_{v3},R,Q,\mathsf{Label}(F_{v3},R,Q))\}$
\EndFor
\end{algorithmic}
\end{algorithm}

\subsection{Preference Pair Templates (DPO Stage 2)}
\label{app:pref-pairs}
For contradiction cases, the preferred reasoning trace $y_w$ correctly detects the inconsistency in the verification step and halts; the rejected trace $y_l$ proceeds with deductive inference.

\textbf{Preferred ($y_w$):}~``Step 1: Detect contradiction between premises. Step 2: Halt reasoning.''\quad
\textbf{Rejected ($y_l$):}~``Step 1: Premises are consistent. Step 2: Continue deductive inference.''

\subsection{Stage 1 Hyperparameters}
\label{app:hyper}
Epochs $3$; LR $2{\times}10^{-5}$; batch $4$; max length $512$; AdamW (mixed precision); LoRA $r{=}8$, $\alpha{=}16$, dropout $0.05$.

\subsection{Ablation Tables}
\label{app:ablation}

\begin{table}[h]
\centering\small
\caption{Ablation: SFT on a vanilla \emph{Mixed-Aug} corpus \emph{without} the conflict-aware verification preamble. Compared to Fusion-LRA (Conflict-Aware SFT, Table~\ref{tab:comparison_results}, Base $=0.988$ on Qwen2), removing the preamble drops Qwen2's base accuracy to $0.5250$, isolating the contribution of the structural prior from the contribution of training data alone.}
\label{tab:stage1_results}
\begin{tabular}{lcccccc}
\toprule
& \multicolumn{2}{c}{\textbf{BERT}} & \multicolumn{2}{c}{\textbf{Qwen2}} & \multicolumn{2}{c}{\textbf{TinyLlama}} \\
\cmidrule(lr){2-3}\cmidrule(lr){4-5}\cmidrule(lr){6-7}
\textbf{Split} & Acc & $\Delta$ & Acc & $\Delta$ & Acc & $\Delta$ \\
\midrule
base                            & 1.0000 & 0.0000 & 0.5250 & -0.4750 & 0.5375 & -0.4625 \\
variant2 (Essential Deletion)   & 0.2500 & -0.7500 & 0.4050 & -0.1200 & 0.5325 & -0.0050 \\
variant3 (Contradiction)        & 0.0000 & -1.0000 & 0.0000 & -1.0000 & 0.0000 & -1.0000 \\
\bottomrule
\end{tabular}
\end{table}

\begin{table}[h]
\centering\small
\caption{Iterative Comparison: Fusion-LRA vs.\ earlier iterations.}
\label{tab:final_performance_comparison}
\begin{tabular}{lcccc}
\toprule
\textbf{Method} & \textbf{Base Acc} & \textbf{Var 2} & \textbf{Var 3} & \textbf{Rank} \\
\midrule
Mixed-Aug              & 0.525          & 0.405          & 0.972          & 3 \\
RA-CoT                 & 0.263          & 0.593          & 0.690          & 2 \\
\midrule
\textbf{Fusion-LRA}    & \textbf{0.988} & \textbf{0.753} & 0.705          & \textbf{1} \\
\bottomrule
\end{tabular}
\end{table}

\subsection{Lean Pilot Study (8 hand-crafted traces)}
\label{app:lean-pilot}

Before the benchmark-scale Lean evaluation reported in Table~\ref{tab:lean_bench}, we ran a small hand-crafted pilot on the Socrates contradiction example to verify (i)~that the Python$\to$Lean bridge faithfully translates a halt-on-contradiction trace into a kernel-checkable theorem, (ii)~that valid halt traces receive reward~$+1$ while opportunistic deductive continuations and malformed traces receive $-1$, and (iii)~that per-trace verification latency is compatible with online RL. Table~\ref{tab:lean_pilot} reports kernel agreement on $8$ representative traces.

\begin{table}[h]
\centering\small
\caption{Lean~4 step-level reward pilot on the Socrates example. Lean kernel agrees with the expected RLVF-Lean reward on $8/8$ traces, including \texttt{sorry}-based admits and syntactic errors. Mean latency $0.36$\,s/trace on CPU.}
\label{tab:lean_pilot}
\begin{tabular}{llrrr}
\toprule
\textbf{Trace ID} & \textbf{Category} & \textbf{Expected $r$} & \textbf{Lean $r$} & \textbf{Time (s)} \\
\midrule
T1-halt-minimal              & correct halt       & $+1$ & $+1$ & 0.36 \\
T2-halt-direct               & correct halt       & $+1$ & $+1$ & 0.38 \\
T3-halt-contradiction-tactic & correct halt       & $+1$ & $+1$ & 0.36 \\
T4-naive-stops-early         & na\"ive continue   & $-1$ & $-1$ & 0.37 \\
T5-naive-wrong-goal          & na\"ive continue   & $-1$ & $-1$ & 0.35 \\
T6-naive-empty (\texttt{sorry}) & na\"ive continue & $-1$ & $-1$ & 0.36 \\
T7-syntax-error              & malformed          & $-1$ & $-1$ & 0.36 \\
T8-misspelled-lemma          & malformed          & $-1$ & $-1$ & 0.36 \\
\midrule
\multicolumn{2}{l}{\textbf{Agreement with oracle}} & \multicolumn{3}{r}{\textbf{8/8 (100\%)}} \\
\bottomrule
\end{tabular}
\end{table}

\subsection{Lean 4 Feasibility Demo for RLVF-Lean (Phase 2)}
\label{app:lean-demo}

The following Lean~4 script encodes the Socrates contradiction example from
Figure~\ref{fig:fusion_structure} as a verified theorem. It accompanies
Section~\ref{sec:phase2} (\emph{Phase 2: Scaling the Verifier with a Formal
Proof Assistant}) and demonstrates that the step-level reward signal defined
by Eq.~(\ref{eq:lean-reward}) can be sourced from the Lean kernel with no
additional model training. The complete source is included in
\texttt{lean\_demo/SocratesContradiction.lean}; a minimal excerpt is shown
below.

\begin{verbatim}
axiom Person   : Type
axiom Socrates : Person
axiom Man      : Person -> Prop
axiom Mortal   : Person -> Prop
axiom men_are_mortal : forall x : Person, Man x -> Mortal x

-- The halt-on-contradiction behavior that RLVF-Lean is rewarded
-- for discovering. Lean accepts the proof iff the policy has
-- genuinely identified both sides of the contradiction.
theorem socrates_contradiction
    (h_man        : Man Socrates)
    (h_not_mortal : Not (Mortal Socrates)) : False := by
  have h_mortal : Mortal Socrates := men_are_mortal Socrates h_man
  exact absurd h_mortal h_not_mortal
\end{verbatim}

A Python bridge (\texttt{lean\_demo/lean\_verifier\_bridge.py}) synthesizes
one such script per policy rollout and returns the kernel's accept/reject
verdict as the step-level reward $r_t$ in Eq.~(\ref{eq:lean-reward}). The
same theorem additionally refutes the naive ``continue deducing'' policy:
under the contradictory premise set, no total proof of an arbitrary query
exists without first deriving \texttt{False}, so any rollout that claims
\texttt{True} receives reward~$-1$. This preserves the conservative
reasoning semantics of Section~\ref{sec:method} at the level of the Lean
kernel itself.

\subsection{Base Example: Complex Dilemma Reasoning Structure}
\label{app:base-example}
We begin with a comprehensive example that establishes the core reasoning pattern, drawing on a structure analogous to the \textit{Paradox of the Court}\footnote{The Paradox of the Court involves a contract between the teacher Protagoras and his student Euathlus, where the student only pays for lessons if he wins a court case. When Protagoras sues Euathlus for the fee, a paradox arises: if Euathlus wins, he owes nothing, but if he loses, he still avoids payment, creating a logical contradiction about the outcome.}, a classic logical dilemma where multiple possible paths lead to the same conclusion, despite their apparent differences.

\begin{itemize}
    \item \textbf{Facts:}
    \begin{itemize}
        \item Anne is green or blue
    \end{itemize}

    \item \textbf{Rules:}
    \begin{itemize}
        \item Rule 1: If someone is green then they are cold. $\forall x \, (\text{Green}(x) \rightarrow \text{Cold}(x))$
        \item Rule 2: If someone is blue then they are cold. $\forall x \, (\text{Blue}(x) \rightarrow \text{Cold}(x))$
        \item Rule 3: If someone is cold then they are rough. $\forall x \, (\text{Cold}(x) \rightarrow \text{Rough}(x))$
        \item Rule 4: If someone is rough then they are young. $\forall x \, (\text{Rough}(x) \rightarrow \text{Young}(x))$
        \item Rule 5: If someone is young then they are cold. $\forall x \, (\text{Young}(x) \rightarrow \text{Cold}(x))$
        \item Rule 6: If someone is young then they are nice. $\forall x \, (\text{Young}(x) \rightarrow \text{Nice}(x))$
    \end{itemize}

    \item \textbf{Questions:}
    \begin{itemize}
        \item Q1: Anne is cold. True/False? [Answer: T]
        \item Q2: Anne is rough. True/False? [Answer: T]
        \item Q3: Anne is young. True/False? [Answer: T]
        \item Q4: Anne is nice. True/False? [Answer: T]
    \end{itemize}
\end{itemize}

The fact ``Anne is green or blue" combined with Rules 1 and 2 creates a classic dilemma: both possibilities lead to the same conclusion. This dilemma reasoning yields  ``Anne is cold." Rule 3 then derives ``Anne is rough" from cold, Rule 4 derives ``Anne is young" from rough, and Rule 6 derives ``Anne is nice" from young. Rule 5 creates a circular reinforcement but doesn't alter the conclusions.

The logical structure can be represented as:
\[
(G_a \lor B_a) \land (G_a \rightarrow C_a) \land (B_a \rightarrow C_a) \vdash C_a
\]

\begin{itemize}
    \item $G_a$: \text{Green}(Anne)
    \item $B_a$: \text{Blue}(Anne)
    \item $C_a$: \text{Cold}(Anne)
\end{itemize}

\subsection{Variation 1: Rule Reduction with Same Conclusions}

This variation demonstrates that removing redundant rules preserves the AI's ability to reach the same conclusions, illustrating that fewer rules can be equally effective.

\begin{itemize}
    \item \textbf{Facts:}
    \begin{itemize}
        \item Anne is green or blue
    \end{itemize}

    \item \textbf{Rules:}
    \begin{itemize}
        \item Rule 1: If someone is green then they are cold. $\forall x \, (\text{Green}(x) \rightarrow \text{Cold}(x))$
        \item Rule 2: If someone is blue then they are cold. $\forall x \, (\text{Blue}(x) \rightarrow \text{Cold}(x))$
        \item Rule 3: If someone is cold then they are rough. $\forall x \, (\text{Cold}(x) \rightarrow \text{Rough}(x))$
        \item Rule 4: If someone is rough then they are young. $\forall x \, (\text{Rough}(x) \rightarrow \text{Young}(x))$
        \item Rule 6: If someone is young then they are nice. $\forall x \, (\text{Young}(x) \rightarrow \text{Nice}(x))$
    \end{itemize}

    \item \textbf{Questions:}
    \begin{itemize}
        \item Q1: Anne is cold. True/False? [Answer: T]
        \item Q2: Anne is rough. True/False? [Answer: T]
        \item Q3: Anne is young. True/False? [Answer: T]
        \item Q4: Anne is nice. True/False? [Answer: T]
    \end{itemize}
\end{itemize}

The reasoning proceeds identically to the base case: the dilemma from Rules 1-2 yields ``Anne is cold," Rule 3 yields rough, Rule 4 yields young, and Rule 6 yields nice. The removal of Rule 5 has no impact on the conclusions, demonstrating its redundancy.

\textbf{Key Insight}: AI systems that recognize this redundancy can simplify their reasoning processes without sacrificing accuracy, embodying the ``less is more" principle.

The simplified logical structure becomes:
\[
(G_a \lor B_a) \land (G_a \rightarrow C_a) \land (B_a \rightarrow C_a) \vdash C_a
\]

\[
C_a \land (C_a \rightarrow R_a) \vdash R_a
\]

\[
R_a \land (R_a \rightarrow Y_a) \vdash Y_a
\]

\[
Y_a \land (Y_a \rightarrow N_a) \vdash N_a
\]

\begin{itemize}
    \item $G_a = \mathrm{Green}(Anne)$
    \item $B_a = \mathrm{Blue}(Anne)$
    \item $C_a = \mathrm{Cold}(Anne)$
    \item $R_a = \mathrm{Rough}(Anne)$
    \item $Y_a = \mathrm{Young}(Anne)$
    \item $N_a = \mathrm{Nice}(Anne)$
\end{itemize}

\subsection{Variation 2: Rule Equivalence with Different Conclusions}

This variation replaces multiple rules with logically equivalent fewer rules, but interestingly leads to different conclusions due to the modified rule interactions.

\begin{itemize}
    \item \textbf{Facts:}
    \begin{itemize}
        \item Anne is green or blue
    \end{itemize}

    \item \textbf{Rules:} (\emph{This variant deliberately combines two perturbations: a logical-equivalence rewrite of Rules~1+2 into a single Rule~A, \textbf{and} the deletion of Rules~5 and~6. Rule~6 is essential for deriving ``nice''; we discuss the two perturbations separately below.})
    \begin{itemize}
        \item Rule A: If someone is green or blue then they are cold. $\forall x \, ((\text{Green}(x) \lor \text{Blue}(x)) \rightarrow \text{Cold}(x))$ \quad \emph{(replaces Rules 1+2)}
        \item Rule 3: If someone is cold then they are rough. $\forall x \, (\text{Cold}(x) \rightarrow \text{Rough}(x))$
        \item Rule 4: If someone is rough then they are young. $\forall x \, (\text{Rough}(x) \rightarrow \text{Young}(x))$
        \item \emph{Rule~5 (Young $\to$ Cold) -- removed (redundant cycle)}
        \item \emph{Rule~6 (Young $\to$ Nice) -- removed (essential for ``nice''; this is the cause of Q4=F, not the Rule~A rewrite)}
    \end{itemize}

    \item \textbf{Questions:}
    \begin{itemize}
        \item Q1: Anne is cold. True/False? [Answer: T]
        \item Q2: Anne is rough. True/False? [Answer: T]
        \item Q3: Anne is young. True/False? [Answer: T]
        \item Q4: Anne is nice. True/False? [Answer: F]
    \end{itemize}
\end{itemize}

Rule~A directly captures the dilemma of Rules 1 and 2, leading to ``Anne is cold" with equivalent logical force---this is the equivalence rewrite, and it preserves Q1--Q3. The change in Q4 (``nice'' becomes F) is caused entirely by the \emph{deletion of Rule~6}, not by the equivalence rewrite. We made both edits in this single illustrative variant to keep the example compact, but they are structurally independent: Variant~4 in the main benchmark applies the rewrite alone, and Variant~2 applies the essential-rule deletion alone.

\textbf{Key Insight}: Logical equivalence of a sub-rule set (here Rules~1+2 vs.\ Rule~A) is a local property; it does not by itself change derivability. Macro-level derivability changes are caused by which essential rules remain, as isolated cleanly in Variant~2.

The logical equivalence can be shown as:
\[
(\forall x \, (G_x \rightarrow C_x) \land \forall x \, (B_x \rightarrow C_x))
\equiv
\forall x \, ((G_x \lor B_x) \rightarrow C_x)
\]

\begin{itemize}
    \item $G_x = \mathrm{Green}(x)$
    \item $B_x = \mathrm{Blue}(x)$
    \item $C_x = \mathrm{Cold}(x)$
\end{itemize}

However, the missing Rule 6 prevents the derivation of Nice(Anne), showing that local equivalence doesn't preserve global derivability.

\subsection{Variation 3: Rule Interference with Contradictory Conclusions}

This variation adds distracting and potentially contradictory rules, testing the AI's ability to address conflicts and maintain coherent reasoning.

\begin{itemize}
    \item \textbf{Facts:}
    \begin{itemize}
        \item Anne is green or blue
        \item Anne is not cold or not nice
    \end{itemize}

    \item \textbf{Rules:}
    \begin{itemize}
        \item Rule 1: If someone is green then they are cold. $\forall x \, (\text{Green}(x) \rightarrow \text{Cold}(x))$
        \item Rule 2: If someone is blue then they are cold. $\forall x \, (\text{Blue}(x) \rightarrow \text{Cold}(x))$
        \item Rule 3: If someone is cold then they are rough. $\forall x \, (\text{Cold}(x) \rightarrow \text{Rough}(x))$
        \item Rule 4: If someone is rough then they are young. $\forall x \, (\text{Rough}(x) \rightarrow \text{Young}(x))$
        \item Rule 5: If someone is young then they are cold. $\forall x \, (\text{Young}(x) \rightarrow \text{Cold}(x))$
        \item Rule 6: If someone is young then they are nice. $\forall x \, (\text{Young}(x) \rightarrow \text{Nice}(x))$

    \end{itemize}

    \item \textbf{Questions:}
    \begin{itemize}
        \item Q1: Anne is cold. True/False? [Answer: F]
        \item Q2: Anne is rough. True/False? [Answer: F]
        \item Q3: Anne is young. True/False? [Answer: F]
        \item Q4: Anne is nice. True/False? [Answer: F]
    \end{itemize}
\end{itemize}

\subsection{Reasoning Process and Analysis}

    This variation introduces \textit{contradictory interference}, which challenges the AI's ability to address conflicts within the logical structure. The fact ``Anne is green or blue," combined with the original chain of reasoning (Rules 1-6), leads to the conclusions that Anne is cold, rough, young, and nice. However, the fact ``Anne is not cold or not nice" introduces a conflict, as it implies that being nice would mean Anne is not cold. This creates a contradiction that different AI systems might address differently:

\paragraph{Contradiction Handling Strategy.}

In the experiments presented in this paper we adopt a \textit{conservative}
contradiction-handling strategy. Once a contradiction is detected in the
premise set, the reasoning process halts and no further deductions are
performed. As a result, all queries associated with that instance are
labeled \texttt{False}.

Alternative reasoning strategies such as priority-based resolution or
paraconsistent reasoning are possible, but they are outside the scope
of the current work and are left for future investigation.

\begin{itemize}
    \item \textbf{Conservative approach}: Detect contradiction and withhold conclusions
    \item \textbf{Priority-based approach}: Apply rule priorities or specificity heuristics
    \item \textbf{Paraconsistent approach}: Accept some contradictions and continue reasoning
\end{itemize}

In this case, a conservative reasoning system would recognize the contradiction and potentially reject all derived conclusions, resulting in \textit{false} for all questions.

 The addition of interfering rules not only tests the AI's ability to ignore distractions but also its capacity for contradiction detection and resolution.

The contradiction can be formally represented as:
\[
(C_a \land N_a) \land (N_a \rightarrow \neg C_a) \vdash \bot
\]

\begin{itemize}
    \item $C_a = \mathrm{Cold}(Anne)$
    \item $N_a = \mathrm{Nice}(Anne)$
    \item $\bot$ represents a value that is always false.
\end{itemize}

\subsection{Comparative Analysis}

Table~\ref{comparison_conclusion} provides a clear comparison of how different modifications to the rule set lead to distinct conclusion patterns, highlighting the sensitivity of reasoning systems to structural changes. These variations demonstrate how even minor adjustments to the logical framework can significantly impact the reasoning process and the final outcomes.

\begin{table}
\centering
\begin{tabular}{|l|c|c|c|c|}
\hline
\textbf{Variations} & \textbf{Cold} & \textbf{Rough} & \textbf{Young} & \textbf{Nice} \\
\hline
Base Example & T & T & T & T \\
Rule Reduction & T & T & T & T \\
Rule Equivalence & T & T & T & F \\
Rule Interference & F & F & F & F \\
\hline
\end{tabular}
\caption{Comparison of conclusions across variations}
\label{comparison_conclusion}
\end{table}

\begin{table}[t]
\centering
\begin{tabular}{lcc|cc|cc}
\hline
\textbf{Split} & \multicolumn{2}{c|}{\textbf{BERT}} & \multicolumn{2}{c|}{\textbf{Qwen2}} & \multicolumn{2}{c}{\textbf{TinyLlama}} \\
 & Acc & $\Delta$ & Acc & $\Delta$ & Acc & $\Delta$ \\
\hline
base                      & 1.0000 & 0.0000  & 1.0000 & 0.0000  & 1.0000 & 0.0000 \\
variant1                  & 1.0000 & 0.0000  & 1.0000 & 0.0000  & 1.0000 & 0.0000 \\
variant2                  & 0.2950 & -0.7050 & 0.2500 & -0.7500 & 0.2500 & -0.7500 \\
variant3                  & 0.0000 & -1.0000 & 0.0000 & -1.0000 & 0.0000 & -1.0000 \\
variant4-contrapositive   & 1.0000 & 0.0000  & 1.0000 & 0.0000  & 1.0000 & 0.0000 \\
variant4-double-negation  & 1.0000 & 0.0000  & 1.0000 & 0.0000  & 1.0000 & 0.0000 \\
variant4-implication      & 1.0000 & 0.0000  & 0.9525 & -0.0475 & 1.0000 & 0.0000 \\
variant4-de-morgan        & 1.0000 & 0.0000  & 1.0000 & 0.0000  & 1.0000 & 0.0000 \\
variant4-identity         & 1.0000 & 0.0000  & 1.0000 & 0.0000  & 1.0000 & 0.0000 \\
variant4-commutativity    & 0.9925 & -0.0075 & 1.0000 & 0.0000  & 1.0000 & 0.0000 \\
variant4-multi (2--5 laws)& 1.0000 & 0.0000  & 0.6450 & -0.3550 & 0.9925 & -0.0075 \\
\hline
\end{tabular}
\caption{Accuracy and deviation from base ($\Delta$) for all models across structural variants.}
\label{tab:structural_variants}
\end{table}

Table~\ref{tab:structural_variants} reports accuracy (Acc) and deviation from the base condition
($\Delta = \mathrm{Acc}_{\text{variant}} - \mathrm{Acc}_{\text{base}}$) for BERT,
Qwen2, and TinyLlama across all structural variants. All models achieve
Acc = 1.0000 on the base split and exhibit no degradation under redundant
rule removal (Variant 1; $\Delta = 0$). By contrast, removing an essential rule
(Variant 2) yields a substantial drop (BERT: 0.2950, $\Delta = -0.7050$;
Qwen2/TinyLlama: 0.2500, $\Delta = -0.7500$), indicating strong sensitivity
to missing inferential links. Injecting explicit contradictory facts
(Variant 3) reduces accuracy to 0.0000 for all models ($\Delta = -1.0000$),
suggesting that the models do not reliably revise conclusions in the presence
of inconsistency.


\end{document}